%% file: main.tex
\documentclass{article}
\usepackage{spconf,amsmath,graphicx,comment,multirow}
\usepackage{enumitem}
\usepackage{tablefootnote}
\usepackage{booktabs}
\input{macros.tex}

\title{A Baseline on Continual Learning Methods for Video Action Recognition}
\name{Giulia Castagnolo$^1$, Concetto Spampinato$^1$,  Francesco Rundo$^2$, Daniela Giordano$^1$, Simone Palazzo$^1$}
\address{$^1$PeRCeiVe Lab, University of Catania, Italy\\$^2$ADG, R\&D Power and Discretes, STMicroelectronics,  Catania, Italy}

\begin{document}
%
\maketitle
\section{Abstract}
\label{sec:abstract}
\input{sections/0-abstract}

\begin{keywords}
Continual Learning, Video Action Recognition, Benchmark, Memory Efficiency
\end{keywords}
\section{Introduction}
\label{sec:introduction}
\input{sections/1-intro}

\section{Related work}
\label{sec:related}
\input{sections/2-related}

\section{Method}
\label{sec:method}
\input{sections/3-method}

\section{Experimental results}
\label{sec:results}
\input{sections/4-results}

\section{Conclusion}
\label{sec:conclusion}
\input{sections/5-conclusions}



\bibliographystyle{IEEEbib}
\bibliography{refs}

\end{document}

%% file: macros.tex
\usepackage[utf8]{inputenc}
\usepackage{soulutf8}
\usepackage{pifont}
\usepackage{amsmath}
\usepackage{amssymb}
\usepackage{color,soul}
\usepackage[dvipsnames]{xcolor}
\usepackage{gensymb}
\usepackage{bm}

\usepackage{xcolor}
\usepackage{hyperref}
\hypersetup{
    colorlinks,
    linkcolor={red!80!black},
    citecolor={blue!50!black},
    urlcolor={blue!80!black}
}

\newcommand{\mytilde}{\raise.17ex\hbox{$\scriptstyle\mathtt{\sim}$}}

\let\emptyset\varnothing
\newcommand{\YesV}{\ding{51}}%

\def\mm{\mathbf{m}}

\def\xx{\mathbf{x}}

\def\XX{\mathbf{X}}

\def\cC{\mathcal{C}}
\def\dD{\mathcal{D}}

\def\lL{\mathcal{L}}
\def\mM{\mathcal{M}}

\def\tT{\mathcal{T}}

%% file: sections/0-abstract.tex
Continual learning has recently attracted attention from the research community, as it aims to solve long-standing limitations of classic supervisedly-trained models.
However, most research on this subject has tackled continual learning in simple image classification scenarios. In this paper, we present a benchmark of state-of-the-art continual learning methods on video action recognition. Besides the increased complexity due to the temporal dimension, the video setting imposes stronger requirements on computing resources for top-performing rehearsal methods.
To counteract the increased memory requirements, we present two method-agnostic variants for rehearsal methods, exploiting measures of either model confidence or data information to select memorable samples. Our experiments show that, as expected from the literature, rehearsal methods outperform other approaches; moreover, the proposed memory-efficient variants are shown to be effective at retaining a certain level of performance with a smaller buffer size.

%% file: sections/1-intro.tex

Video action recognition is a well-known problem in computer vision, given to its wide range of applications, including surveillance~\cite{videosurveillance}, behavior understanding~\cite{activitynet}, content-based retrieval~\cite{videoretrival}. It is also a complex problem, due to the difficulty to learn spatio-temporal action patterns in a high-dimensional space. Recent advances in deep learning, including attention-based vision transformers~\cite{vit,viVit,spacetimeattention,recurringtransformer}, have achieved unprecedented results on video action recognition benchmarks, pushing performance of automatic methods closer to humans'.

However, current approaches for supervised video action recognition assume a \emph{stationary} data distribution, where all dataset classes and samples are simultaneously available at training time. Recently, this hypothesis has been put under question, giving rise to a line of research on \emph{continual learning} methods, designed to deal with changes in the class distribution (new classes may become available, while others are replaced) and/or in the intra-class data distribution. The relaxation of the stationarity assumption introduces several novel challenges; notably, models trained on sequences of tasks tend to focus on currently-available classes, degrading performance on previously-seen classes --- a phenomenon known as ``catastrophic forgetting''; similarly, in presence of gradual shifts in the data distribution of a certain class, models become unable to correctly process ``older'' samples.

While several solutions to continual learning have been proposed, based on architectural priors~\cite{pnn}, knowledge distillation~\cite{lwf}, regularization~\cite{ewc} or experience replay~\cite{er}, they have been mostly validated on simple image datasets, such as MNIST, CIFAR10/100 or simplified versions of ImageNet. Hence, it is hard to assess their suitability to more complex use cases, such as video action recognition.

This work intends to assess whether state-of-the-art methods for continual learning generalize to high-dimensional spatio-temporal data. To achieve this goal, we define an experimental protocol for continual learning in video action recognition, by selecting multiple class subsets from UCF-101~\cite{ucf101} and evaluating the performance of state-of-the-art approaches at the task. We also propose two simple method-agnostic modifications, namely, \emph{confidence-driven rehearsal} and \emph{information-driven downsampling}, which define criteria for the selection of samples for experience replay, in the attempt to build a more effective buffer for retaining past knowledge with fewer samples.

Results show that the best performance is achieved by combining experience replay with knowledge distillation from past samples, e.g., DER~\cite{der} and DER++~\cite{der}. Additionally, we show that the proposed memory-efficient variants are able to significantly improve the performance of even the best methods in our benchmark, providing an efficient way to improve performance in such complex scenario.

%% file: sections/2-related.tex
A continual learning problem is typically divided into a sequence of \emph{tasks}, presented one at a time. In classification, each task may require training a model on a new set of classes or on different data distributions of previously-seen classes. 
Continual learning aims to cope with catastrophic forgetting in this scenario, using different techniques. 

\emph{Rehearsal} methods keep a \emph{buffer} of samples from previous tasks, to prevent the model from forgetting past knowledge. Methods in this category include DER/DER++~\cite{der}, ER~\cite{er}, GSS~\cite{gss}, FDR~\cite{fdr}, HAL~\cite{hal}, GEM~\cite{gem}, AGEM \cite{agem} and iCarl~\cite{icarl}. 
\emph{Knowledge distillation} approaches employ an earlier version of the model (trained on previous tasks) to transfer features or to encourage the current model to emulate past predictions. The pioneering work in this category is LwF~\cite{lwf}; recent approaches have also attempted to integrate auxiliary knowledge from unrelated tasks~\cite{boschini2022transfer}.
\emph{Regularization} methods, e.g., EWC~\cite{ewc}, introduce loss terms to counteract modification of backbone features in favor of new tasks. Regularization and knowledge distillation are often employed alongside other techniques: DER/DER++~\cite{der} regularize a cross-entropy loss by enforcing logit similarity of buffered past samples.
\emph{Architectural} methods, such as PackNet~\cite{packnet} and HAT~\cite{hat}, progressively extend a model's backbone to cope with new tasks, but require that model capacity must increase with the number of tasks. Pruning methods~\cite{clnp,piggyback} may help mitigate this issue.

To date, rehearsal methods show the most promising results. Unfortunately, the presence of a buffer and the corresponding memory overhead may become significant, if not prohibitive, when dealing with video sequences, because of the increase introduced by the temporal dimensions.

Finally, it should be mentioned that a related problem, i.e., \emph{class-incremental learning}, has already been studied in video action recognition~\cite{park2021class}. However, class-incremental learning --- i.e., progressively showing new classes to a classification model --- only analyzes a portion of the problem (for instance, it does not address \emph{task-incremental} performance) and generally applies a pre-training on a large part (often, half~\cite{park2021class,douillard2020podnet,hou2019learning}) of the original dataset, which is not a common procedure in the literature for continual learning.

%% file: sections/3-method.tex
\subsection{Problem formulation}

In this work, as in most of continual learning literature, we focus on classification tasks. Hence, let \emph{task} $\tT\left( \cC \right) \sim p(\tT)$ be a classification problem defined on a set of classes $\cC = \left\{ y_1, \dots, y_{c}\right\}$. Given two tasks $\tT_i$ and $\tT_j$, we assume that the corresponding sets of classes are different, i.e., $\cC_i \cap \cC_j = \emptyset$: in this context, we define the problem as of either \emph{task-incremental learning} (T-IL) or \emph{class-incremental learning} (C-IL), based on whether knowledge of the task to which a sample belongs is provided to the model at inference time or not, respectively. 

A \emph{continual learning problem} consists in a sequence of $T$ tasks $\left( \tT_1, \dots, \tT_T \right)$ sampled from $p(\tT)$. A model $\mM$ is allowed to train on each task $\tT_i$, before moving to task $\tT_{i+1}$; at a generic task $\tT_{i}$, data from any other tasks cannot be accessed (unless previously store by the model, as in the case of rehearsal methods).
Given the task sequence $\left( \tT_1, \dots, \tT_T \right)$, model $\mM$, parameterized by $\bm{\theta}$, is trained to optimize a classification objective $\lL$ (commonly, a cross-entropy loss) on each task at a time, while attempting to prevent performance decreases on previous tasks.

\subsection{Evaluation procedure}
\label{sec:eval}

Given a \emph{source dataset} $\dD_s$, including videos for a set of class labels $\cC_s$, we emulate a sampling of the $p(\tT)$ distribution by splitting the set of classes into random groups of $c$ classes each. As a result, we obtain a set of tasks $\mathbb{T} = \left\{ \tT_1, \tT_2, \dots, \tT_N \right\}$, with $\tT_i$ representing a portion of $\dD_s$.

We can sample a \emph{continual learning problem} $\left\{ \tT_1, \dots, \right.$ $\left. \tT_T \right\}$ by selecting a random subset of $T$ tasks from $\mathbb{T}$, and train each of the methods under analysis on that problem, to guarantee a fair comparison. This procedure is then repeated for $E$ experiments. For each experiment, a fraction $p_\text{test}$ of samples from each class is left out as a test set.

Evaluation metrics, averaged over the set of $E$ experiments, include the following:
\begin{itemize}[noitemsep,nolistsep]
\item Accuracy in task-incremental learning: for each task, we compute test performance using only predictions for the classes included in that task, and average the computed accuracy scores over the set of tasks.
\item Accuracy in class-incremental learning: for each task, we compute test performance using predictions for all classes, and average the computed accuracy scores over the set of tasks.
\item Buffer size: for rehearsal methods, number of elements stored in the buffer and required memory space.
\end{itemize}

\subsection{Memory-efficient variants}

Our results (see Sect.~\ref{sec:results}) show that, as expected from the literature, rehearsal methods significantly outperform other paradigms. However, buffer memory requirements increase significantly for video sequences. We hereby propose two model-agnostic variants for buffer management, aimed at reducing the number of elements to store while preserving classification accuracy.\\
\textbf{Confidence-driven rehearsal.} High-dimensionality of videos reflects on a lower generalization power by a set of random buffered samples, due to the curse of dimensionality.
Hence, rather than attempting to model the entire distributions of a task's classes, an alternative lies in selecting samples on which the model is \emph{most confidently correct}: while this may hinder the recognition of under-represented class modes, it reinforces knowledge on the most discriminant portion of the distribution, which is also expected to cover the most density mass.
In practice, a sample $(\xx, y)$ is eligible for buffering if the model's prediction for class $y$ is above a threshold $\delta$.\\
\textbf{Information-driven downsampling.} Common video classification models require a fixed input size, making it is necessary to downsample longer sequences. Usually, such downsampling is agnostic to frame content; hence, memory efficiency may be achieved through frame selection, so that more informative content is stored in a smaller buffer. We propose to employ the norm of optical flow~\cite{farneback2003two} as a measure of importance, as it quantifies motion and redundancy between consecutive frames.
Formally, given a video sequence $\XX = \left\{ \xx_1, \dots, \xx_p \right\}$, we compute motion vectors $\left\{ \mm_1, \dots, \mm_{p-1} \right\}$, such that $\mm_i = \text{OF}(\mm_i, \mm_{i+1})$, where $\text{OF}$ is the optical flow function; then, we select the subset of frames with the largest $L_2$ norm $\left\| \mm_i \right\|_2$, ordered by their original position in the video.

%% file: sections/4-results.tex
\subsection{Methods}

Our benchmark includes state-of-the-art continual learning methods, covering the range of paradigms from the literature: DER~\cite{der}, DER++~\cite{der}, ER~\cite{er}, FDR~\cite{fdr}, HAL~\cite{hal}, GSS~\cite{gss}, GEM~\cite{gem}, AGEM~\cite{agem}, EWC~\cite{ewc} and LwF~\cite{lwf}\footnote{We excluded iCarl~\cite{icarl} due to excessive memory requirements, as the reference implementation required a concatenation of all samples of a task with the current buffer.}. For a fair comparison, all methods employ the same backbone network, i.e., R(2+1)D~\cite{r2plus1d}: we excluded architectural methods~\cite{packnet,hat} from our analysis, as model capacity extension would lead to an unfair comparison. All methods included in our experiments are agnostic to data modality, and can be applied to videos without any modifications. Implementations are taken from the Mammoth continual learning library\footnote{\url{https://github.com/aimagelab/mammoth}}.

\subsection{Dataset and task definition}
\label{sec:dataset}

As source dataset $\dD_s$, we employ UCF101~\cite{ucf101}, a video action recognition dataset featuring 101 categories, with approximately 100-150 videos each; video duration is mostly between 2 and 10 seconds, although longer videos may be present for certain categories. In our experiments, we employ a subset $\cC_s$ of 30 classes, by selecting distinguishable actions, removing too similar classes and classes with less than 130 videos. Classes are grouped into pairs (consistently with continual learning procedures on other datasets, e.g., CIFAR10) to create the task set $\mathbb{T}$, including $N=15$ different tasks. Then, we define a fixed set of 50 continual learning problems, each being a task sequence $\left\{ \tT_1, \dots, \tT_T \right\}$ of $T=5$ tasks randomly sampled from $\mathbb{T}$. To ensure class balance, for each class we select a subset of 130 videos, with 100 videos used as a training set and the rest as a test set ($p_\text{test} = 3/10$).

\subsection{Training details}

We normalize each color channel to zero mean and unitary standard deviation, and resize spatially to 160$\times$160 pixels. Data augmentation includes random cropping at 128$\times$128, random horizontal flipping, and temporal cropping by selecting 16 consecutive frames from a random point. At test time, we apply center-cropping and process all non-overlapping 16-frame windows from an entire video.
The R(2+1)D backbone is trained from scratch for 80 epochs per task, using the RMSProp optimizer (learning rate: $10^{-5}$, batch size: 16); in our preliminary experiments, this configuration empirically yielded the most stable results, compared to standard SGD (which showed slow convergence) or Adam (which, for some methods, led to exploding losses). Method-specific hyperparameters were set to default values from the original papers. 
Training was carried out on a machine with 8-core Intel Xeon Skylake CPU, 64 GB RAM, NVIDIA V100 GPU.

\begin{table}[!h]
\centering
\caption{Classification accuracy in class-incremental learning (C-IL) and task-incremental learning (T-IL). Rehearsal methods are marked with a \YesV.}
\label{tab:comparison_sota}
\begin{tabular}{lccccc}
    \toprule
        \textbf{Model} & \textbf{Rehars.} & \textbf{C-IL} & \textbf{T-IL} \\
        \midrule
        DER++~\cite{der}  & \YesV & $47.87\pm4.31$ & $90.60\pm2.21$\\
        DER~\cite{der} & \YesV & $39.80\pm3.25$ & $88.80\pm3.64$ \\
        ER~\cite{er}  & \YesV & $31.80\pm6.02$ & $90.21\pm1.84$ \\ 
        FDR~\cite{fdr}   & \YesV & $29.33\pm3.46$ &  $80.87\pm3.12$ \\ 
        HAL~\cite{hal}   & \YesV & $25.61\pm 6.35$ & $69.40\pm8.26$ \\ 
        GSS~\cite{gss}   & \YesV & $20.60\pm2.97$ &  $71.86\pm3.76$ \\ 
        AGEM~\cite{agem} & \YesV & $18.53\pm2.70$ &  $83.40\pm1.46$ \\ 
        LwF~\cite{lwf}   & & $17.13\pm 0.99$ & $57.00\pm4.14$ \\
        GEM~\cite{gem}   & \YesV & $16.13\pm4.36$ &  $72.33\pm9.00$ \\ 
        EWC~\cite{ewc}   & & $11.87\pm1.71$ &  $50.33\pm5.20$ \\ 
    \bottomrule
    \end{tabular}
\end{table}

\begin{table}[!h]
\centering
\caption{Impact of the proposed \emph{confidence-driven rehearsal} (CDR) variant on classification accuracy, for different buffer sizes and values of the $\delta$ threshold.}
\label{tab:cdr}
\begin{tabular}{lccccc}
\toprule
 & \textbf{Buffer} & $\bm{\delta}$ & \textbf{C-IL} & \textbf{T-IL} \\
\midrule
\multirow{9}{*}{\rotatebox{90}{DER++~\cite{der}}} & \multirow{4}{*}{500 (2.2 GB)} & -     & $48.20\pm2.41$ & $91.90\pm1.74$ \\
                                       &                                & $0.6$ & $47.22\pm1.39$ & $91.25\pm2.21$ \\
                                       &                                & $0.7$ & $47.00\pm3.21$ & $92.00\pm3.09$ \\
                                       &                                & $0.8$ & $49.22\pm4.40$ & $90.89\pm2.11$ \\
                                       \cmidrule(l){2-5}
                                       & \multirow{4}{*}{200 (0.9 GB)}  & -     & $47.87\pm4.31$ & $90.60\pm2.21$ \\
                                       &                                & $0.6$ & $46.47\pm4.46$ & $89.07\pm4.39$ \\
                                       &                                & $0.7$ & $47.80\pm5.07$ & $90.20\pm3.54$ \\
                                       &                                & $0.8$ & $49.40\pm4.92$ & $88.47\pm2.19$ \\
                                       \cmidrule(l){2-5}
                                       & \multirow{4}{*}{100 (0.5 GB)}  & -     &  
                                       $42.00\pm2.96$ & $88.33\pm3.90$ \\
                                       &                                & $0.6$ & $41.00\pm4.20$ & $86.25\pm2.87$ \\
                                       &                                & $0.7$ & $40.11\pm2.50$ & $87.00\pm0.58$ \\
                                       &                                & $0.8$ & $40.00\pm4.15$ & $88.22\pm4.70$ \\
\midrule
\multirow{9}{*}{\rotatebox{90}{DER~\cite{der}}}   & \multirow{4}{*}{500 (2.2 GB)}       & -      & $44.93\pm4.88$ & $90.00\pm1.11$ \\
                                       &                                & $0.6$ &  $43.40\pm5.15$ & $90.86\pm3.11$ \\
                                       &                                & $0.7$ & $43.60\pm2.60$ & $91.33\pm1.61$ \\
                                      &                                 & $0.8$ & $43.65\pm5.21$ & $87.33\pm3.79$ \\
                                       \cmidrule(l){2-5}
                                       & \multirow{4}{*}{200 (0.9 GB)}  & -     & $39.80\pm3.25$ & $88.80\pm3.64$ \\
                                       &                                & $0.6$ & $40.80\pm5.31$ & $85.87\pm3.34$ \\
                                       &                                & $0.7$ & $44.20\pm4.65$ & $87.53\pm3.05$ \\
                                       &                                & $0.8$ & $44.33\pm5.10$ & $88.73\pm2.30$ \\
                                       \cmidrule(l){2-5}
                                       & \multirow{4}{*}{100 (0.5 GB)}  & -     &  $33.27\pm3.80$ & $84.95\pm6.71$ \\
                                       &                                & $0.6$ & 
                                       $35.33\pm5.25$ & $84.67\pm5.60$ \\  &                             & $0.7$ &  $36.73\pm3.47$ & $88.26\pm6.52$ \\
                                       &                                & $0.8$ & 
                                       $37.67\pm5.17$ & $86.53\pm3.62$ \\
    \bottomrule
    \end{tabular}
\end{table}

\subsection{Results}
We first report the results obtained by state-of-the-art methods under comparison. For rehearsal methods, we employ a buffer size of 200 (a standard value from the literature), corresponding --- based on our pre-processing --- to an increase in memory requirements by 940 MB. Table~\ref{tab:comparison_sota} shows each method's performance in terms of class-incremental and task-incremental classification accuracy; we report mean and standard deviations over the set of continual learning problems defined by our experimental protocol. Methods employing experience replay perform significantly better, as expected from previous literature results on images. It is interesting to note that the top three methods, i.e., DER++, DER and ER achieve very similar task-incremental accuracy, showing effectiveness in learning each individual task, but differ significantly in the capability to retain previous knowledge.
In absolute terms, DER++ yields promising accuracy (47.87\% in class-incremental learning), considering that its performance on images (CIFAR-10) is about 65\%~\cite{der}, with the same buffer size. 

We then assess the impact of the proposed memory-efficient variants on the best-performing methods only, i.e., DER and DER++. Table~\ref{tab:cdr} shows the effect of the proposed \emph{confidence-driven rehearsal} (CDR) technique, for different buffer sizes and values of the $\delta$ confidence threshold. In the case of DER, enabling CDR with a 100-element buffer pushes class-incremental performance closer to those with buffer size 200; similarly, CDR with a 200-element buffer reaches similar performance as achieved with a 500-element buffer. On the other hand, CDR with DER does not have a significant impact when applied to a 500-element buffer, which may be large enough to compensate for confidence improvements. CDR does not seem to improve task-incremental accuracy, which is reasonable, since it acts on the buffer and mainly addresses forgetting. Applying CDR on DER++ is less effective, though some improvements can be seen, especially with buffer size of 200. However, it is reasonable to assume that the stronger recovery capabilities of DER++ make the contribution of CDR less important.
For both DER and DER++, a general trend shows that larger $\delta$ values lead to better performance; however, we found that exceeding the $\delta=0.8$ threshold harms performance: we hypothesize that the distribution of selected samples excessively narrows the data distribution, worsening generalization.
We then evaluate how our \emph{information-driven downsampling} (IDD), based on optical flow, affects model accuracy. Table~\ref{tab:idd} shows that enabling IDD has a positive impact on class-incremental accuracy on DER with small buffers, while it is less effective with larger buffer. On DER++, IDD positively affects only the usage of a 100-element buffer, possibly for similar reasons as discussed in the case of CDR. However, in no case IDD is able to compensate for a smaller buffer size, showing a superiority by CDR in this respect.

\begin{table}[!h]
\centering
\caption{Impact of the proposed \emph{information-driven downsampling} (IDD) variant on the performance of DER and DER++, for different buffer sizes.}
\label{tab:idd}
\begin{tabular}{lccccc}
\toprule
 & \textbf{Buffer} & \textbf{IDD} & \textbf{C-IL} & \textbf{T-IL} \\
\midrule
\multirow{7}{*}{\rotatebox{90}{DER++~\cite{der}}} & \multirow{2}{*}{500 (2.2 GB)} &      & $48.20\pm2.41$ & $90.90\pm1.74$ \\
                                       &  &  \YesV     &  $48.10\pm1.80$ & $89.66\pm2.01$\\
                                       \cmidrule(l){2-5}
                                       & \multirow{2}{*}{200 (0.9 GB)}  &       & $47.87\pm4.31$ & $89.60\pm2.21$ \\
                                       &  & \YesV & $46.00\pm1.11$ & $89.93\pm0.76$ \\
                                       \cmidrule(l){2-5}
                                       & \multirow{2}{*}{100 (0.5 GB)}  &      &  
                                       $42.00\pm2.96$ & $88.33\pm3.90$ \\ 
                                       &  & \YesV & $44.64\pm3.64$ & $90.13\pm2.50$\\
\midrule
\multirow{7}{*}{\rotatebox{90}{DER~\cite{der}}}   & \multirow{2}{*}{500 (2.2 GB)}        &        & $44.93\pm4.88$ & $89.00\pm1.11$ \\ &   & \YesV & $42.50\pm3.82$ & $88.75\pm3.11$ \\
                                
                                       \cmidrule(l){2-5}
                                       & \multirow{2}{*}{200 (0.9 GB)}  &       & $39.80\pm3.25$ & $88.80\pm3.64$ \\
                                       &   & \YesV & $42.20\pm3.58$ & $87.33\pm3.81$ \\
                                       \cmidrule(l){2-5}
                                       & \multirow{2}{*}{100 (0.5 GB)}  &      &  
                                       $33.27\pm3.80$ & $84.95\pm6.71$ \\                       &        & \YesV & $34.93\pm4.12$ & $84.33\pm1.66$ \\
    \bottomrule
    \end{tabular}
\end{table}

%% file: sections/5-conclusions.tex
We presented a benchmark of state-of-the-art continual learning methods for video action recognition, showing that methods designed for images are able, to a certain extent, to generalize to videos, achieving promising performance in class-incremental and task-incremental settings. We also propose two memory-efficient variants for buffer sample selection, demonstrating that the CDR variant helps to retain (or even improve) performance even when reducing the buffer size, while the IDD variant is less effective in this regard.

Future improvements of this work will address two main research directions to improve the proposed memory-efficient variants. First, we mean to explore advances to the proposed \emph{confidence-driven rehearsal}, by integrating mechanisms for automatic and adaptive threshold setting. Second, rather than explicitly defining a measure for \emph{information-driven downsampling} (e.g., optical flow norm), we intend to investigate the employment of attention mechanisms to find measures of correlation between input and model representations, thus using the latter as a reference for importance estimation.